\documentclass[10pt,twocolumn,letterpaper]{article}

\usepackage[pagenumbers]{cvpr} %
\usepackage[accsupp]{axessibility}

\usepackage[pdftex]{graphicx}
\usepackage{amsmath}
\usepackage{amssymb}
\usepackage{booktabs}
\usepackage{multirow}

\newcommand{\remove}[1]{}
\newcommand{\ilker}[1]{}
\newcommand{\dy}[1]{}
\newcommand{\aerd}[1]{}

\newcommand{\segmentationTask}[0]{Referring Expression Segmentation}
\newcommand{\localizationTask}[0]{Referring Expression Comprehension}
\newcommand{\colorizationTask}[0]{Language-guided Image Colorization}
\newcommand{\shortSegmentationTask}[0]{RES}
\newcommand{\shortLocalizationTask}[0]{REC}
\newcommand{\shortColorizationTask}[0]{LIC}

\usepackage[pagebackref,breaklinks,colorlinks]{hyperref}

\usepackage[capitalize]{cleveref}
\crefname{section}{Sec.}{Secs.}
\Crefname{section}{Section}{Sections}
\Crefname{table}{Table}{Tables}
\crefname{table}{Tab.}{Tabs.}

\def\cvprPaperID{16} %
\def\confName{MULA}
\def\confYear{2022}

\begin{document}

\title{Modulating Bottom-Up and Top-Down Visual Processing via Language-Conditional Filters}

\author{Ilker Kesen\textsuperscript{1,2}
\quad Ozan Arkan Can\textsuperscript{3}
\quad Erkut Erdem\textsuperscript{1,4}
\quad Aykut Erdem\textsuperscript{1,2}
\quad Deniz Yuret\textsuperscript{1,2}\vspace{0.2cm}\\
\textsuperscript{1} Ko\c{c} University, KUIS AI Center~ \textsuperscript{2} Ko\c{c} University, Computer Engineering Department \\
\textsuperscript{3} Amazon Search~ \textsuperscript{4} Hacettepe University, Computer Engineering Department~
}

\maketitle
\begin{abstract}
How to best integrate linguistic and perceptual processing in multi-modal tasks that involve language and vision is an important open problem.
In this work, we argue that the common practice of using language in a top-down manner, to direct visual attention over high-level visual features, may not be optimal. We hypothesize that the use of language to also condition the bottom-up processing from pixels to high-level features can provide benefits to the overall performance.
To support our claim, we propose a U-Net-based model and perform experiments on two language-vision dense-prediction tasks: \MakeLowercase{\segmentationTask} and \MakeLowercase{\colorizationTask}. We compare results where either one or both of the top-down and bottom-up visual branches are conditioned on language. Our experiments reveal that using language to control the filters for bottom-up visual processing in addition to top-down attention leads to better results on both tasks and achieves competitive performance. Our linguistic analysis suggests that bottom-up conditioning improves segmentation of objects especially when input text refers to low-level visual concepts. Code is available at \url{https://github.com/ilkerkesen/bvpr}.
\end{abstract}

\section{Introduction}
\label{sec:intro}
As human beings, we can easily perceive our surroundings  with our visual system and interact with each other using language. Since the work of Winograd \cite{winograd_understanding_1972}, developing a system that understands human language in a situated environment has been one of the long-standing goals of artificial intelligence.
Recent successes of deep learning studies in both language and vision domains have increased the interest in tasks that combine language and vision \cite{antol_vqa_2015, xu_show_2015,krishna_visual_2016,suhr_corpus_2017,anderson_vision-and-language_2018,hudson_gqa_2019}. However, how to best integrate linguistic and perceptual processing is still an important open problem. Towards this end, we investigate whether language should be used for conditioning bottom-up visual processing as well as top-down attention.

In the human visual system, attention is driven by both \emph{top-down} cognitive processes (\emph{e.g.} focusing on a given shape) and \emph{bottom-up} salient, behaviourally relevant stimuli (\textit{e.g.} fast moving objects and contrasting colors) \cite{corbetta_control_2002,connor_visual_2004,theeuwes_topdown_2010}. Studies on embodied language explore the link between linguistic and perceptual representations \cite{pulvermuller_words_1999,vigliocco_representing_2004,gallese_brains_2005} and often assume that language has a \textit{high-level} effect on perception and drives the \emph{top-down} visual attention \cite{bloom_how_2002,jackendoff_foundations_2002,dessalegn_more_2008}. However, recent studies from cognitive science point out that language comprehension also affects low-level visual processing \cite{meteyard_motion_2007,boutonnet_words_2015,lupyan_words_2015}. Motivated by this, we propose a model that can modulate either or both of \emph{bottom-up} and \emph{top-down} visual processing with language and compare different designs for language modulation.

 Current deep learning systems for language-vision tasks typically start with low-level image processing, then connect the language representation with high-level visual features to control the visual focus. To integrate both modalities, concatenation \cite{malinowski_ask_2015}, element-wise multiplication \cite{lu_hierarchical_2016,kim_multimodal_2016}, attention from language to vision \cite{xu_ask_2016,yang_stacked_2016,lu_knowing_2017,anderson_bottom-up_2018,zellers_recognition_2019} and transformers \cite{tan2019lxmert,Deng_2021_ICCV,vision-language-transformer} are commonly used. These studies typically do not condition low-level visual features on language. Some methods \cite{de_vries_modulating_2017,perez_film_2018} do the opposite by conditioning only the bottom-up visual processing on language.

 To evaluate language-modulation on the bottom-up and top-down visual branches independently, we develop an architecture that clearly separates these two branches (based on U-Net \cite{ronneberger_u-net_2015}) and allows us to experiment with modulating one or both branches with language. %
 The bottom-up branch starts from low-level visual features and applies a sequence of contracting filters that result in successively higher level feature maps with lower spatial resolution. Following this, a top-down branch takes the final low resolution feature map and applies a sequence of expanding filters that eventually result in a map with the original image resolution. Information flows between branches through skip connections between contracting and expanding filters at the same level. Our proposed architecture is task-agnostic and it can be used for various vision-language tasks involving dense prediction. We evaluate our model with different language-modulation settings on two different tasks: \MakeLowercase{\segmentationTask} and 
 \MakeLowercase{\colorizationTask}. % 

In the \textit{\MakeLowercase{\segmentationTask}} (\shortSegmentationTask) task, given an image and a natural language description, the aim is to obtain the segmentation mask that marks the object(s) described. We can contrast this with pure image based object detection \cite{girshick_fast_2015,ren_faster_2017} and semantic segmentation \cite{long_fully_2015,chen_deeplab_2017} tasks which are limited to predefined semantic classes. %
The language input may contain various visual attributes (e.g. \textit{shape}), spatial information (e.g. \textit{in front of}), actions (e.g. \textit{running}) and interactions/relations between different objects (e.g. \textit{arm of the chair that the cat is sitting on}).

In \textit{\MakeLowercase{\colorizationTask}} (\shortColorizationTask) task, given a grayscale image and a description, the aim is to predict pixel color values. The absence of color information in the input images makes this problem interesting to experiment with because color words do not help in conditioning the bottom-up branch when the input image is grayscale. %

We find that conditioning both branches leads to better results, achieving competitive performance on both tasks.
Our experiments suggest that conditioning the bottom-up branch on language is important to ground low-level visual information.
On {\shortSegmentationTask}, we find that modulating only the bottom-up branch performs significantly better than modulating only the top-down branch especially when color-dependent language is present in the input.
Our findings on \shortColorizationTask~show that when color information absent in input images, the bottom-up baseline naturally fails to predict and manipulate colors of target objects specified by input language.  That said, conditioning the bottom-up branch still improves the colorization quality by helping our model to accurately segment and colorize the target objects as a whole.  %

The rest of the paper is structured as follows: We summarize related work and compare it to our approach in Section \ref{sec:related}. We describe our model in detail in Section \ref{sec:segmentation-model}. We share the details of our experiments in Section \ref{sec:experiments}. Section \ref{sec:conclusion} summarizes our contributions.

\section{Related Work}
\label{sec:related}

\noindent \textbf{\localizationTask~(\shortLocalizationTask)}.
In this problem, the goal is to locate a bounding box for the object(s) described in the input language. The proposed solutions can be divided into two categories: two-stage and one-stage methods. Two-stage methods \cite{mao_generation_2016,hu_modeling_2017,nagaraja_modeling_2016,yu_mattnet_2018,wang_neighbourhood_2019,cirik_using_2018,hu_modeling_2017,yu_mattnet_2018,liu_learning_2019,yang2020graph-structured} rely on a pre-trained object detector \cite{ren_faster_2017,he_mask_2017} to generate object proposals in the first stage. In the second stage, they assign scores to the object proposals depending on how much they match with input language. One-stage methods \cite{luo2020multi,yang2020propagating,Liao_2020_CVPR,yang2019fast,yang2020improving,Deng_2021_ICCV} directly localize the referred objects in one step. Most of these methods condition only the top-down visual processing on language, while some fuse language with multi-level visual representations.  %

\noindent \textbf{\segmentationTask~(\shortSegmentationTask)}.
In this task, the aim is to generate a segmentation mask for the object(s) referred in the input language \cite{hu_segmentation_2016}. To accomplish this, multi-modal LSTMs \cite{liu_recurrent_2017,margffoy-tuay_dynamic_2018}, ConvLSTMs \cite{shi_convolutional_2015,liu_recurrent_2017,chen_see-through-text_2019,ye2020dual}, word-level attention \cite{shi_key-word-aware_2018,hui_linguistic_2020,chen_see-through-text_2019,ye2020dual}, cross-modal attention \cite{ye_cross-modal_2019,hu_bi-directional_2020,huang2020referring,luo2020multi,luo2020cascade,jing2021locate}, and transformers~\cite{vaswani_attention_2017,vision-language-transformer} have been used. Each one of these methods modulates only the top-down branch with language. As one exception, EFN \cite{feng2021encoder} conditions the bottom-up branch on language, but does not modulate the top-down branch with language.

\noindent \textbf{\colorizationTask~(\shortColorizationTask)}. In this task, the aim is to predict colors for all the pixels of a given input grayscale image based on input descriptive text.
Specifically, \cite{manjunatha_learning_2018} inserts extra Feature-wise Linear Modulation (FiLM) layers \cite{perez_film_2018} into a pre-trained ResNet to predict color values in LAB color space.
Multi-modal LSTMs \cite{liu_recurrent_2017,zou_language-based_2019} and generative adversarial networks \cite{goodfellow_generative_2014,bahng_coloring_2018,chen_language-based_2018} are also used in this context to colorize sketches. Similar to us, Tag2Pix \cite{kim_tag2pix_2019} extends U-Net to perform colorization on line art data, but it modulates only the top-down visual processing with symbolic input using concatenation.  %

\noindent \textbf{Language-conditional Parameters}.
Here we review methods that use input-text-dependent dynamic parameters to process visual features. To control a visual model with language, MODERN and FiLM \cite{de_vries_modulating_2017,perez_film_2018} used conditional batch normalization layers with language-conditioned coefficients rather than customized filters. Numerous methods \cite{li_tracking_2017,gao_question-guided_2018,gavrilyuk_actor_2018,margffoy-tuay_dynamic_2018,chen_referring_2019,misra_mapping_2018} generate language-conditional dynamic filters to convolve visual features. Some \shortSegmentationTask~models \cite{margffoy-tuay_dynamic_2018,chen_referring_2019} also incorporate language-conditional filters into the their top-down visual processing.
To map instructions to actions in virtual environments, LingUNet \cite{misra_mapping_2018} extends U-Net by adding language-conditional filters to the top-down visual processing only.
Each one of these methods conditions either top-down or bottom-up branch only.

\noindent \textbf{Comparison.} To support our main research question, our architecture clearly separates bottom-up and top-down visual processing. This allows us to experiment with modulating either one branch or both branches with language and evaluate their individual contributions. The majority of related work conditions only the top-down visual processing on language. Other U-Net-based methods \cite{misra_mapping_2018,kim_tag2pix_2019} and most transformer models \cite{vision-language-transformer,Deng_2021_ICCV} which implement cross-modal attention between textual and visual representations in top-down visual processing fall into this category. A few exceptions \cite{de_vries_modulating_2017,perez_film_2018,feng2021encoder} do the opposite by conditioning only the bottom-up branch.  Some methods \cite{luo2020multi,yang2019fast,huang2020referring} fuse language with a multi-level visual representation, which leads to good results, but this kind of fusion does not allow the evaluation of language conditioning on top-down vs bottom-up visual processing. Our architecture allows language to control either or both of top-down and bottom-up branches. We show that (i) the bottom-up conditioning is important to ground language to low-level visual features, and (ii) conditioning both branches on language leads to best results.

\begin{figure*}[!t]
\centering
\includegraphics[width=0.85\linewidth]{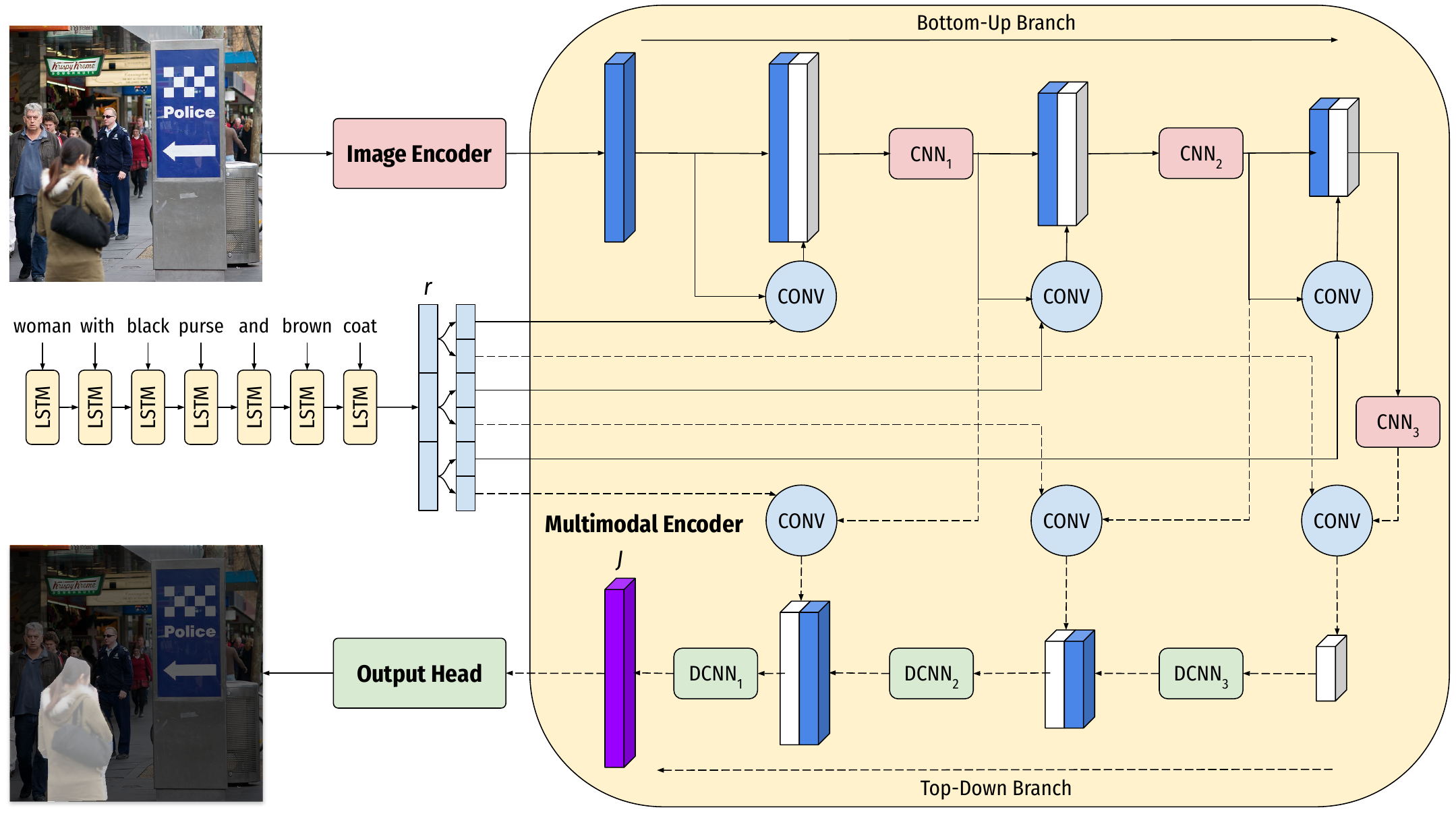}
\caption{Overview of the proposed model on the task of \MakeLowercase{\segmentationTask}. } 
\label{fig:architecture}
\end{figure*}

\section{The Model}
\label{sec:segmentation-model}
Here, we describe our proposed model in detail. \figurename~\ref{fig:architecture} shows an overview of our proposed architecture. First, the model extracts a 
tensor of low-level features using a pre-trained convolutional neural network and encodes the given natural language expression
to a vector representation
using an LSTM \cite{hochreiter_long_1997}. Starting with the visual feature tensor, the model generates feature maps through a contracting and expanding path where the output head takes the final map and performs dense prediction, similar to U-Net \cite{ronneberger_u-net_2015}. Our proposed architecture modulates both of the contracting and expanding paths with language using convolutional filters generated from the given expression. It is important to emphasize that the previous works either have a language-guided top-down or a language-conditional bottom-up visual processing branch. As will be discussed in the next section, our experiments show that modulating both of these paths improves the performance dramatically.

\subsection{Low-level Image Features}
Given an input image $I$, we extract visual features $I_0$ by using a pre-trained convolutional network. In the \shortSegmentationTask~task, we use DeepLab-v3+ backbones \cite{chen2018encoder}, and in the \shortColorizationTask~task, we use ResNet-101 pre-trained on ImageNet \cite{he_deep_2016,deng_imagenet_2009}. On the task of \MakeLowercase{\segmentationTask}, we also concatenate 8-D location features to this feature map following the previous work \cite{liu_recurrent_2017,ye_cross-modal_2019}.  %

\subsection{Language Representation}
Consider a language input $S = [w_1, w_2, ... , w_n]$ where each word $w_i$ is represented with a $300$-dimensional GloVe embedding \cite{pennington_glove_2014}. We map the language input $S$ to hidden states using an LSTM as $h_i = \mbox{LSTM}(h_{i-1}, w_i)$. We use the final hidden state of the LSTM as the language representation, $r = h_n$. Later on, we split this language representation into pieces to generate language-conditional filters. %

\subsection{Multi-modal Encoder}
After generating image 
and language representations, 
our model generates a multi-modal feature map  representing the input image and the given natural language expression.
Our multi-modal encoder module extends U-Net by conditioning both contracting and expanding branches on language using language-conditional filters.

In the bottom-up branch, our model applies $m$ convolutional modules to the image representation $I_0$. Each module, $\mbox{CNN}_i$, takes the concatenation of the previously generated feature map ($\textit{F}_{i-1}$) and its convolved version with language-conditional filters $K_{i}^{F}$ and produces an output feature map ($F_{i}$). Each $\mbox{CNN}_i$ has a 2D convolution layer followed by batch normalization \cite{ioffe_batch_2015} and ReLU activation function \cite{maas_rectifier_2013}. The convolution layers have $5\times5$ filters with $stride=2$ and $padding=2$ halving the spatial resolution, and they all have the same number of output channels.

Similar to \cite{misra_mapping_2018}, we split the textual representation $r$ to $m$ equal parts ($t_i$), and then use each $t_i$ to generate a language-conditional filter for $i$th bottom-up layer ($K_{i}^{F}$):
\begin{equation}
K_{i}^{F} = \mbox{AFFINE}_{i}^{F}(t_i)
\end{equation}

Each $\mbox{AFFINE}_{i}^{F}$ is an affine transformation followed by normalizing and reshaping to convert the output to convolutional filters.
After obtaining the filters, we convolve them over the feature map obtained from the previous module ($F_{i-1}$) to relate expressions to image features:
\begin{equation}
G_{i}^{F} = \mbox{CONVOLVE}(K_{i}^{F}, F_{i-1})
\end{equation}

Then, the concatenation of these text-modulated features ($G_{i}^{F}$) for $i$th bottom-up layer and the previously generated features ($F_{i-1}$) is fed into module $\mbox{CNN}_i$ for the next step.

In the top-down branch, we generate $m$ feature maps starting from the final output of the contracting branch as:
\begin{align}
G_{i}^{H}&{}={}\mbox{CONVOLVE}(K_{i}^{H}, F_{i})\\
H_{m}&{}={}\mbox{DCNN}_i(G_{m}^{H})\\
H_{i}&{}={}\mbox{DCNN}_i(G_{i}^{H} \oplus H_{i-1})
\end{align}

Similar to the bottom-up branch, $G_{i}^{H}$ is the modulated feature map with language-conditional filters defined as:
\begin{equation}
K_{i}^{H} = \mbox{AFFINE}_{i}^{H}(t_i)
\end{equation}

where $\mbox{AFFINE}_i^{H}$ is again an affine transformation followed by normalizing and reshaping for $i$th layer of the top-down branch. Here, we convolve the filter ($K_{i}^{H}$) over the feature maps from the contracting branch ($F_{i}$). Each upsampling module $\mbox{DCNN}_i$ gets the concatenation ($\oplus$) of the text related features and the feature map ($H_{i}$) generated from the previous module. Only the first module operates on just convolved features. Each $\mbox{DCNN}_i$ consists of a 2D deconvolution layer followed by a batch normalization and ReLU activation function. Final output $H_1$ becomes our joint feature map $J$ representing the input image / language pair. The deconvolution layers have $5\times5$ filters with $stride=2$ and $padding=2$ doubling the spatial resolution, and they all have the same number of output channels.

\subsection{Output Heads}
As mentioned earlier, we develop our model as a generic solution which can be used to solve language-vision problems involving dense prediction. In this direction, we adapt our model to two different tasks by varying the output head: \MakeLowercase{\segmentationTask} and \MakeLowercase{\colorizationTask}.

\noindent \textbf{Segmentation.} In the \MakeLowercase{\segmentationTask} problem, the goal is to generate segmentation mask for a given input image and language pair. After generating the joint feature map $J$, we apply a stack of layers ($D_1$, $D_2$, ..., $D_m$) to map $J$ to the exact image size. Similar to upsampling modules, each $D_k$ is a 2D deconvolution layer followed by batch normalization and the ReLU activation. Each $D_k$ preserves the number of channels except for the last one which maps the features to a single channel for the mask prediction. We omit the batch normalization and the ReLU activation for the final module, instead we apply a sigmoid function to turn the final features into probabilities. Given these probabilities and ground-truth mask, we train our network by using binary cross entropy loss.

\noindent \textbf{Colorization}. In the \MakeLowercase{\colorizationTask} task, the goal is to predict pixel color values for given input image with the guidance of language input. A convolutional layer by with 3$\times$3 filters generates class scores for each spatial location of $J$. We apply bilinear upsampling to these predicted scores to match input image size. Given predicted scores and ground-truth color classes, we train our model by using a weighted cross entropy loss. To create compound LAB color classes and their weights, we follow the exactly same process with \cite{manjunatha_learning_2018,zhang_colorful_2016}.

\begin{table*}
\centering
\begin{tabular}{l|ccc|ccc|ccc|c}
\hline
\multirow{2}{4em}{Method} & \multicolumn{3}{c|}{UNC} & \multicolumn{3}{c|}{UNC+} & \multicolumn{3}{c|}{G-Ref} & ReferIt \\ \cline{2-11}
 & val & testA & testB & val & testA & testB & val (G) & val (U) & test (U) & test \\ \hline \hline
CMSA \cite{ye_cross-modal_2019} & 58.32 & 60.61 & 55.09 & 43.76 & 47.60 & 37.89 & 39.98 & - & - & 63.80 \\
STEP \cite{chen_see-through-text_2019} & 60.04 & 63.46 & 57.97 & 48.19 & 52.33 & 40.41 & 46.40 & - & - & 64.13 \\
BRINet \cite{hu_bi-directional_2020} & 61.35 & 63.37 & 59.57 & 48.57 & 52.87 & 42.13 & 48.04 & - & - & 63.46 \\
CMPC \cite{huang2020referring} & 61.36 & 64.53 & 59.64 & 49.56 & 53.44 & 43.23 & 49.05 & - & - & 65.53 \\
LSCM \cite{hui_linguistic_2020} & 61.47 & 64.99 & 59.55 & 49.34 & 53.12 & 43.50 & 48.05 & - & - & 66.57 \\
EFN \cite{feng2021encoder} & 62.76 & 65.69 & 59.67 & 51.50 & 55.24 & 43.01 & \textbf{51.93} & - & - & \textbf{66.70} \\ 
BUSNet \cite{yang2021bottom} & 63.27 & 66.41 & \textbf{61.39} & \textbf{51.76} & \textbf{56.87} & \textbf{44.13} & 50.56 & - & - & - \\
\hline
Our model & \textbf{64.63} & \textbf{67.76} & 61.03 & \textbf{51.76} & 56.77 & 43.80 & 50.88 & \textbf{52.12} & \textbf{52.94} & 66.01 \\ \hline \hline
MCN\textsuperscript{\textdagger} \cite{luo2020multi} & 62.44 & 64.20 & 59.71 & 50.62 & 54.99 & 44.69 & - & 49.22 & 49.40 & - \\
CGAN\textsuperscript{\textdagger} \cite{luo2020cascade} & 64.86 & 68.04 & 62.07 & 51.03 & 55.51 & 44.06 & 46.54 & 51.01 & 51.69 & - \\
LTS\textsuperscript{\textdagger} \cite{jing2021locate} & 65.43 & 67.76 & 63.08 & 54.21 & 58.32 & 48.02 & - & 54.40 & 54.25 & - \\
VLT\textsuperscript{\textdagger} \cite{vision-language-transformer} & 65.65 & 68.29 & 62.73 & \textbf{55.50} & 59.20 & \textbf{49.36} & 49.76 & 52.99 & \textbf{56.65} & - \\
\hline
Our model\textsuperscript{\textdagger} & \textbf{67.01} & \textbf{69.63} & \textbf{63.45} & 55.34 & \textbf{60.72} & 47.11 & \textbf{53.51} & \textbf{55.09} & 55.31 & \textbf{57.09} \\
\hline
\end{tabular}
\caption{Comparison with the previous work by using the overall \textit{IoU} metric. \textdagger~ denotes the corresponding method uses the mean \textit{IoU} metric. "-" indicates that the model has not been evaluated on that dataset.}
\label{table:refexp-mainresults}
\end{table*}

\section{Experimental Analysis}\label{sec:experiments}
This section contains the details of our experiments on \MakeLowercase{\segmentationTask} (Section \ref{sec:refexp-experiments}) and \MakeLowercase{\colorizationTask} (Section \ref{sec:colors-experiments}) tasks\footnote{We provide the implementation details, and present the complete ablation experiments in the supplementary material.}.

\subsection{\segmentationTask}
\label{sec:refexp-experiments}

\noindent \textbf{Datasets.} We evaluate our model on ReferIt (130.5k expressions, 19.9k images) \cite{kazemzadeh_referitgame_2014}, UNC (142k expressions, 20k images), UNC+ (141.5k expressions, 20k images) \cite{yu_modeling_2016} and Google-Ref (G-Ref) (104.5k expressions, 26.7k images) \cite{mao_generation_2016} datasets. Unlike UNC, location-specific expressions are excluded in UNC+ through enforcing annotators to describe objects by their appearance. ReferIt, UNC, UNC+ datasets are collected through a two-player game and have short expressions (avg. 4 words). G-Ref have longer and richer expressions, its expressions are collected from Amazon Mechanical Turk instead of a two-player game. G-Ref does not contain a test split, and Nagaraja et al. \cite{nagaraja_modeling_2016} extends it by having separate splits for validation and test, which are denoted as val (U) and test (U).

\noindent\textbf{Evaluation Metrics.} Following previous work, we use intersection-over-union (\textit{IoU}) and $p@X$ as evaluation metrics. Given the predicted segmentation mask and the ground truth, the \textit{IoU} metric is the ratio between the intersection and the union of the two. There are two different ways to calculate \textit{IoU}: the overall \textit{IoU} calculates the total intersection over total union score throughout the entire dataset and the mean \textit{IoU} calculates the mean of \textit{IoU} scores of each individual example. For a fair comparison, we use both \textit{IoU} metrics. The second metric, $p@X$, calculates the percentage of the examples that have \textit{IoU} score higher than the threshold $X$. %

\noindent\textbf{Quantitative Results.} Table \ref{table:refexp-mainresults} shows the comparison of our model with previous methods. Bold faces highlight the highest achieved scores. We evaluate our model by using both \textit{IoU} metrics for a fair evaluation.
Table \ref{table:refexp-prec} presents the comparison of our model with the state-of-the-art in terms of $p@X$. The difference between our model and the state-of-the-art increases as the threshold increases. This indicates that our model is better at segmenting the referred objects including smaller ones.  %

\begin{table}[!t]
\centering
\resizebox{\linewidth}{!}{
\begin{tabular}{l|rrrrr}
\hline
Method & \textit{p@0.5} & \textit{p@0.6} & \textit{p@0.7} & \textit{p@0.8} & \textit{p@0.9}  \\ \hline \hline
CMSA & 66.44 & 59.70 & 50.77 & 35.52 & 10.96 \\
STEP & 70.15 & 63.37 & 53.15 & 36.53 & 10.45 \\
BRINet & 71.83 & 65.05 & 55.64 & 39.36 & 11.21 \\
LSCM & 70.84 &  63.82 & 53.67 & 38.69 & 12.06 \\
EFN & 73.95 & 69.58 & 62.59 & 49.61 & 20.63 \\
MCN & 76.60 & 70.33 & 58.39 & 33.68 & 5.26 \\
LTS & 75.16 & 69.51 & 60.74 & 45.17 & 14.41 \\
\hline
Our Model & \textbf{76.67} & \textbf{71.77} & \textbf{64.76} & \textbf{51.69} & \textbf{22.73} \\ \hline
\end{tabular}
}
\caption{Comparison with the previous works on the val set of UNC dataset with \textit{p@X} metrics.}
\label{table:refexp-prec}
\end{table}

\begin{figure*}[!t]
\centering
\includegraphics[width=0.95\linewidth]{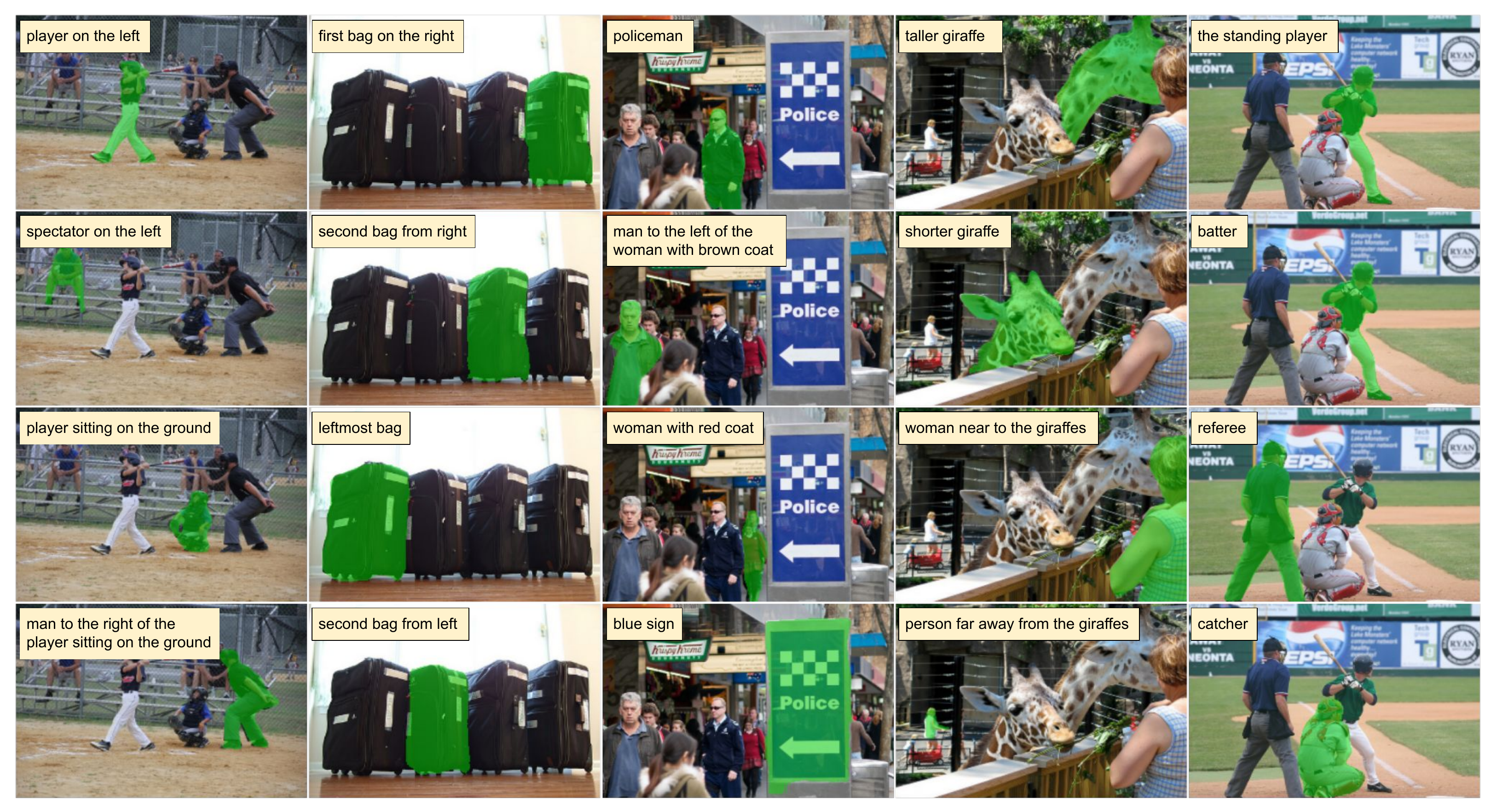}
\caption{Some predictions of our model on the UNC validation set. %
}
\label{fig:positiveExamples}
\end{figure*}

\noindent{\textbf{Qualitative Results.}}
We visualize some of the segmentation predictions of our model to gain better insights about the trained model. \figurename~\ref{fig:positiveExamples} shows some of the cases that our model segments correctly. These examples demonstrate that our model can learn a variety of language and visual reasoning patterns. For example, the first two examples of the fourth column show that our model learns to relate superlative adjectives (e.g., \textit{taller}) with visual comparison. Examples including spatial prepositions (e.g., \textit{near to}) demonstrate the spatial reasoning ability of the model. Our model can also learn a domain-specific nomenclature (e.g. \textit{catcher}) that is present in the dataset. Lastly, we can see that the model can also detect certain actions (e.g., \textit{sitting}).

\begin{figure*}
\centering
\includegraphics[width=\linewidth]{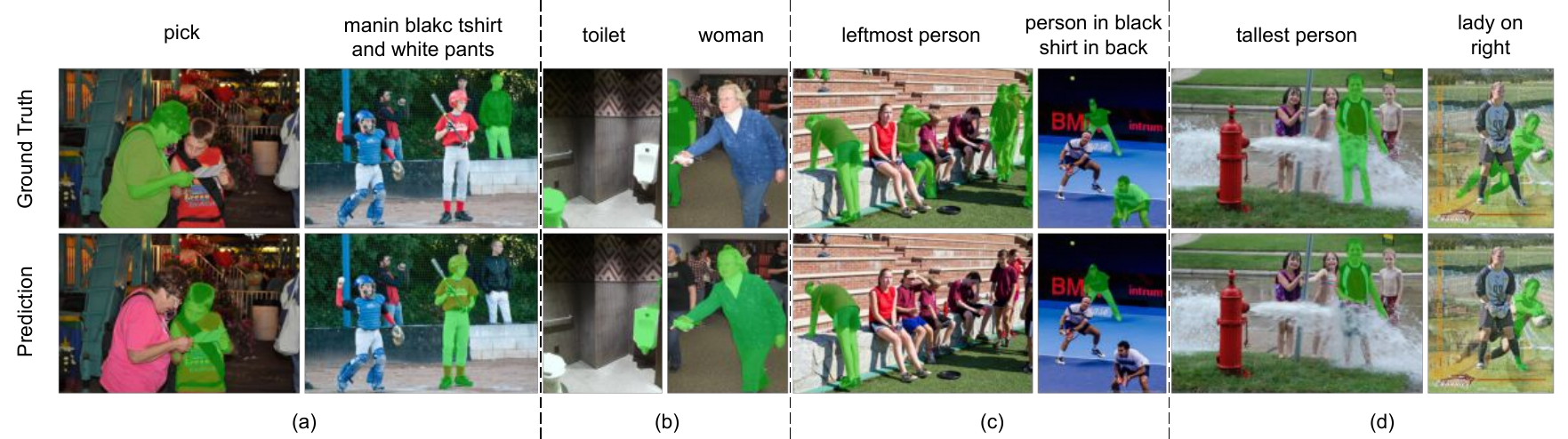}
\caption{Some incorrect predictions of our model on the UNC validation set. Each group (a-d) shows one pattern we observed within the predictions. The first row shows the ground truth mask, the second one is the prediction of our model.}
\label{fig:negativeExamples}
\end{figure*}

\figurename~\ref{fig:negativeExamples} shows failure cases from our model on the UNC test split. Examples in (a) show that our model tends to fail in case of typos. Our model segments the correct objects for these two examples when the typos are fixed (e.g. \textit{pink} instead of \textit{pick}). Examples in (b) show that some of the expressions are ambiguous, where there are multiple objects that could be referred to by the expression. In this case, the model seems to segment the most salient object. Some of the annotations contain incorrect or incomplete ground-truth mask (c). Finally, some of the examples (d) are hard to segment completely due to the lack of light or objects that mask the referred objects.

\begin{table}[!t]
\centering
\resizebox{0.8\linewidth}{!}{

\begin{tabular}{l|l|c}
\hline
Method & Backbone & \textit{IoU} \\ \hline \hline
Top-down Baseline & ResNet-50 & 58.06 \\
Bottom-up Baseline & ResNet-50 & 60.74 \\
Our Model & ResNet-50 & 63.59 \\
Our Model & ResNet-101 & \textbf{64.63} \\
\hline
\end{tabular}}
\caption{Ablation study on the validation set of the UNC dataset with overall \textit{IoU} metric.}
\label{table:ablation}
\end{table}

\noindent{\textbf{Ablation Study.}} We implemented 3 different models, the top-down baseline, the bottom-up baseline and our model, to show the effect of modulating language in expanding and contracting visual branches. While the bottom-up baseline modulates language in bottom-up branch only, the top-down baseline modulates language in top-down branch only. Our model conditions language on both branches. Table \ref{table:ablation} shows us the results. The bottom-up baseline outperforms the top-down one with $\approx$2.7 IoU improvement. Modulating language in both branches yields the best results by improving the bottom-up baseline with $\approx$2.85 IoU score.

\begin{figure}[!t]
\centering
\includegraphics[width=\linewidth]{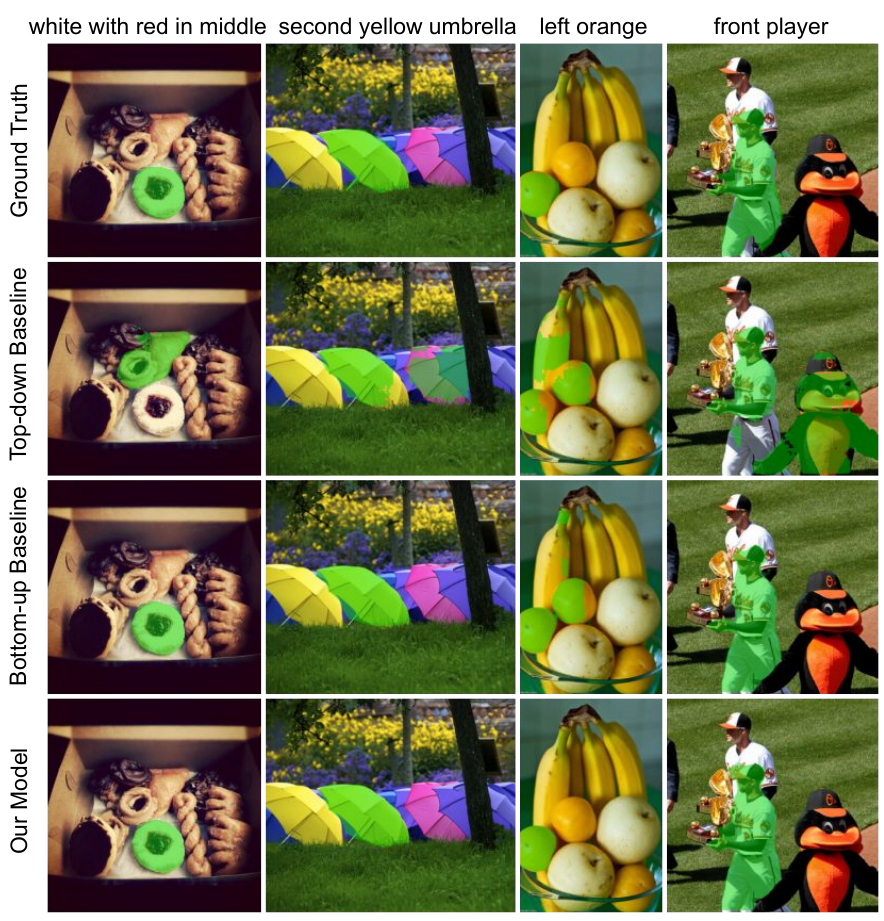}
\caption{Comparison of different architectural setups on the UNC test samples. 
}
\label{fig:branches}
\end{figure}

\figurename~\ref{fig:branches} visualizes the predictions of the different models on the same examples. The bottom-up baseline performs better when the description has color information as we show in the first three examples. The top-down-only baseline also fails to detect object categories in some cases, and segments additional unwanted objects with similar category or appearance (e.g. \textit{banana} vs. \textit{orange}). Overall, our model which conditions both visual branches on language gives the best results.

\begin{table*}[!t]
\centering
\begin{tabular}{  l | c c c c c c } 
 \hline
 Method & +C & -C & +IN & +IN,-C & +JJ* & +JJ*,-C \\
 \hline
 Top-down Baseline & 52.59 & 59.59 & 54.84 & 55.66 & 57.66 & 61.16 \\ 
 Bottom-up Baseline & 60.40 & 60.02 & 56.05 & 55.49 & 61.18 & 61.86 \\ 
 Our Model & 62.98 & 63.57 & 59.60 & 59.27 & 64.45 & 65.56 \\
 \hline
\end{tabular}
\caption{IoU performance of different setups depending on the input expression category for the UNC test splits.}
\label{table:refexp-lang}
\end{table*}

\noindent{\textbf{Language-oriented Analysis.}} To analyze the effect of language on model performance, we divided UNC test splits into subsets depending on the different types of words (e.g. colors) and phrases (e.g. noun phrases with multiple adjectives) included in input expressions. Table \ref{table:refexp-lang} shows us the results of different models on these subsets. The first column stands for models, and the rest stand for different input expression categories. We exclude the categories which do not contribute to our analysis. We use DeepLab-v3+ ResNet-50 as visual backbone in each method. The notation of the categories are similar to Part of Speech (POS) tags \cite{marcus1993building}, where we denote prepositions with IN, examples with adjectives with JJ*, and colors with C. Preceding plus and minus signs stand for inclusion and exclusion. For instance, +IN,-C column stands for the subset where each expression contains at least one preposition without any color words. Color words (e.g. red, darker) has the most impact on the performance in comparison to other types of words and phrases. Our model and the bottom-up baseline performs significantly better than top-down baseline on the subset that includes colors. In the opposite case, where input expressions with colors are excluded, the top-down baseline has performance similar to the bottom-up baseline, and our final model outperforms both single branch models. Since colors can be seen as low-level sensory information, low performance in the absence of the bottom-up branch can be expected. This demonstrates the importance of conditioning the bottom-up visual branch on language to capture low-level visual concepts.

\subsection{\colorizationTask}
\label{sec:colors-experiments}

\noindent{\textbf{Datasets.}} Following the prior work \cite{manjunatha_learning_2018}, we use a modified version of COCO dataset \cite{lin_microsoft_2014} where the descriptions that do not contain color words are excluded. In this modified version, the training split has 66165, and the validation split has 32900 image / description pairs, and all images have a resolution of $224\times224$ pixels.

\noindent \textbf{Evaluation Metrics.} Following the previous work, we use pixel-level top-1 (acc@1) and top-5 accuracies (acc@5) in LAB color space, and additionally PSNR and LPIPS \cite{zhang2018perceptual} in RGB for evaluation. A lower score is better for LPIPS, and a higher score is better for the rest.

\begin{table}
\centering
\resizebox{\linewidth}{!}{
\begin{tabular}{lcccc}
\hline
Method & acc@1 & acc@5 & PSNR & LPIPS \\ \hline
FiLMed ResNet & 23.70 & 60.50 & - & - \\
FiLMed ResNet (ours) & 20.22 & 49.57 & 20.89 & 0.1280 \\
Top-down Baseline & 22.83 & 51.85 & 21.29 & \textbf{0.1226} \\
Bottom-up Baseline & 21.85 & 51.34 & 20.98 & 0.1448 \\
Our Model & 23.38 & 54.27 & 21.42 & 0.1262 \\
Our Model w/o balancing & \textbf{33.74} & \textbf{67.83} & \textbf{22.75} & 0.1250 \\
\hline
\end{tabular}
}
\caption{Colorization results on the modified COCO validation split.}
\label{table:colors-mainresults}
\end{table}

\noindent{\textbf{Quantitative Results.}} We present the quantitative performance of our model in Table \ref{table:colors-mainresults}, and compare it with different design choices and previous work. FiLMed ResNet \cite{manjunatha_learning_2018} uses FiLM \cite{perez_film_2018} to perform language-conditional colorization. FiLMed ResNet (ours) denotes the results reproduced by the implementation provided by the authors. To show the effect of language modulation on different branches, we train 3 different models again: the top-down baseline, the bottom-up baseline and our model. We also re-train our model without class rebalancing and denote it as \textit{Our Model w/o balancing}.

Contrary to the segmentation experiments, the top-down baseline performs better than the bottom-up baseline on the colorization task in all measures. Since, color information is absent in input images, bottom-up branch cannot encode low-level image features by modulating color-dependent language.

When we disable class rebalancing in the training phase, we observe a large improvement in acc@1 and acc@5 due to the imbalanced color distribution, where the model predicts the frequent colors exist in the backgrounds.

\begin{figure}[!t]
\centering
\includegraphics[width=\linewidth]{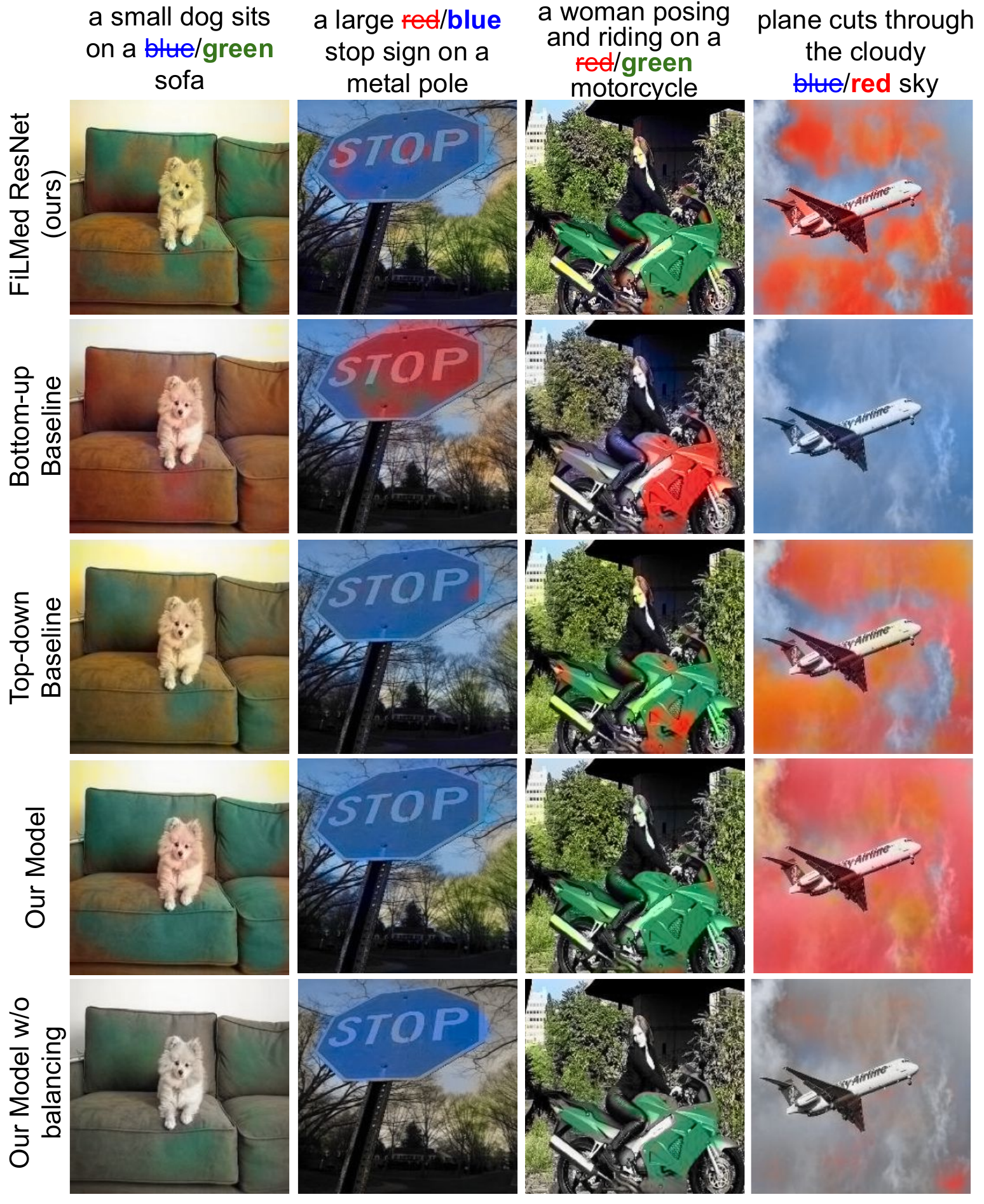}
\caption{Color manipulation performance of different models on COCO validation examples. Each column focuses on a different color conversion.
}
\label{fig:colors-manipulation}
\end{figure}

\noindent{\textbf{Qualitative Results.}}
We visualize some of the colorization outputs of the trained models to analyze them in more detail in \figurename~\ref{fig:colors-manipulation}.
FiLMed ResNet (ours) can understand all colorization hints, and it can manipulate object colors with some incorrectly predicted areas. The top-down baseline also performs similar to FiLMed ResNet (ours), where both models condition only the top-down branch on language.

In this task, since the models are blind to color, the bottom-up baseline loses its effectiveness to some degree, and starts to predict the most probable colors. This can be seen on the second and the last example, where the bottom-up baseline predicts \textcolor{red}{red} for the stop sign and \textcolor{blue}{blue} for the sky. Although, the bottom-up baseline performs worse in this task, modulating the bottom-up branch with language still contributes to our final model to localize and recognize the objects present in the scene. This can be seen on the last two examples, where the top-down baseline mixes colors up in some object parts (e.g. the red parts in the motorcycle). Our model w/o balancing tends to predict more grayish colors (e.g. dog, sky).

\begin{figure}[t]
\centering
\includegraphics[width=\linewidth]{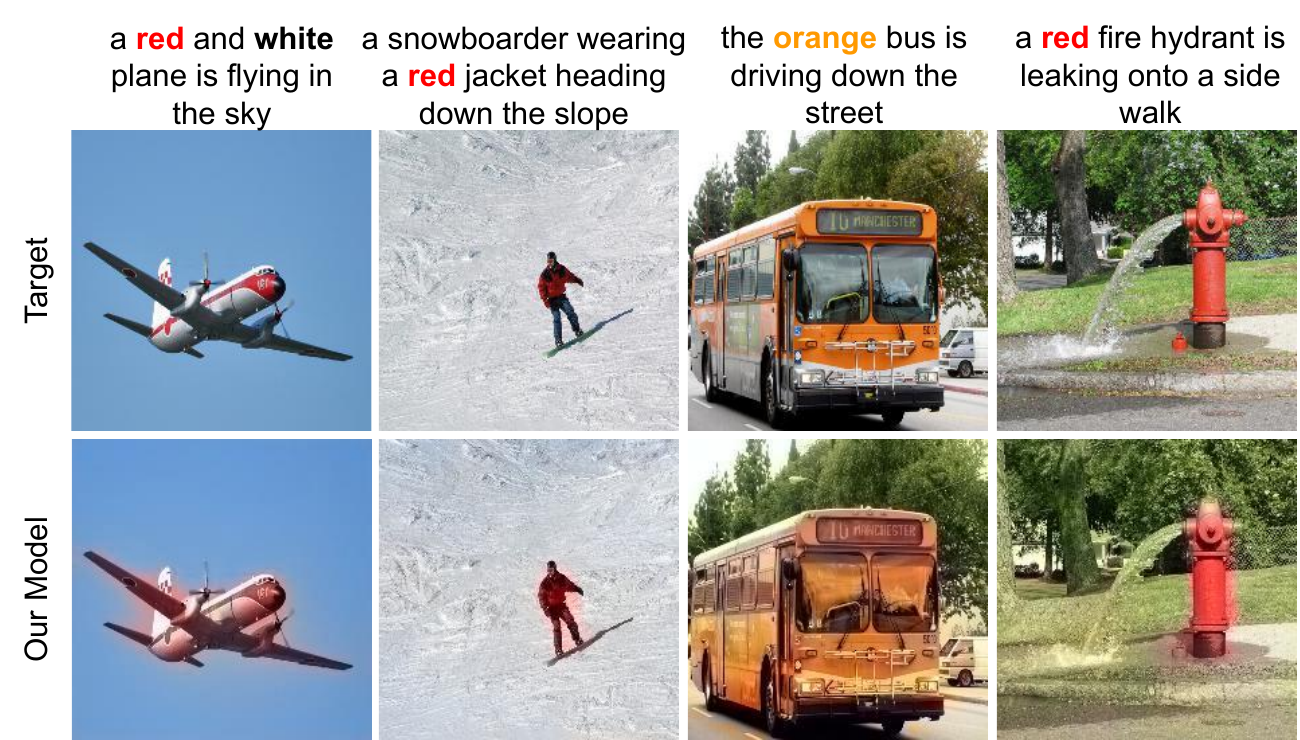}
\caption{Failure cases of our model on the language-guided image colorization task.}
\label{fig:colors-failure}
\end{figure}

\figurename~\ref{fig:colors-failure} highlights some of the failure cases we observed throughout the dataset. In the first two examples, our model is able to localize and recognize the target objects, but it fails to colorize them successfully by colorizing not only the targeted parts but also other parts. Models generally fail to colorize small objects since the data is imbalanced and it contains vast backgrounds and big objects frequently. The last two examples show that models fail to colorize reflective or transparent objects like glasses or water, these were also difficult in the language based segmentation task (see \figurename~\ref{fig:negativeExamples} (d)).

 \section{Conclusion}
\label{sec:conclusion}
In this work, we suggested that conditioning both top-down and bottom-up visual processing on language is beneficial for grounding language to vision. To support this claim, we proposed a generic architecture with explicit bottom-up and top-down visual branches for vision-language problems involving dense prediction. Our experiments on two different tasks demonstrated that conditioning both visual branches on language gives the best results. Our experiments on the \MakeLowercase{\segmentationTask} task revealed that conditioning the bottom-up branch on language plays a vital role to process color-dependent input language. The \MakeLowercase{\colorizationTask} experiments demonstrated similar conclusions, the bottom-up baseline failed to colorize the target objects since the color information is absent in the input images.\\

\noindent \textbf{Limitations.} We share common failure cases in \figurename~\ref{fig:negativeExamples} and \figurename~\ref{fig:colors-failure}. The performance of our model on both tasks decreases in the presence of transparent and/or reflective objects. Our model also fails to colorize small objects, mostly due to having an imbalanced color distribution. Finally, our current model is limited to integrated vision and language tasks involving dense prediction, and we did not perform experiments on other vision and language problems.

\noindent \textbf{Acknowledgements}.
This work was supported in part by an AI Fellowship to I. Kesen provided by the KUIS AI Center, GEBIP 2018 Award of the Turkish Academy of Sciences to E. Erdem, and BAGEP 2021 Award of the Science Academy
to A. Erdem.

{\small
\bibliographystyle{ieee_fullname}
\bibliography{egbib}
}

\clearpage
\appendix
\section{Supplementary Material}
\setcounter{table}{0}
\setcounter{figure}{0}
\renewcommand{\thetable}{A\arabic{table}}
\renewcommand{\thefigure}{A\arabic{figure}}

This supplementary material contains the implementation details (Section \ref{sec:implementation-details}) and the complete ablation studies (Section \ref{sec:complete-ablations}) of our work.

\subsection{Implementation Details}
\label{sec:implementation-details}
\noindent \textbf{\segmentationTask.}
Following previous work \cite{liu_recurrent_2017,margffoy-tuay_dynamic_2018,ye_cross-modal_2019,chen_see-through-text_2019}, we limit the maximum length of expressions to $20$. We set input image size to $512\times512$ and $640\times640$ for training and inference phase respectively.
We use the first four layers of DeepLabv3+ with ResNet-101 backbone, pre-trained on COCO dataset by excluding images appear on the validation and the test sets of UNC, UNC+ and G-Ref datasets similar to  previous work \cite{luo2020multi,vision-language-transformer,yu_mattnet_2018}. Thus, our low-level visual feature map $I$ has the size of $64\times64\times64\times1032$ in training, and $80\times80\times1032$ in inference phase, both including 8-D location features. In all convolutional layers, we set the filter size, stride, and number of filters as $(5,5)$, $2$, and $512$, respectively. The depth is $4$ in the multimodal encoder part of the network. For the bottom-up-only baseline, we used grouped convolution in the bottom-up branch to prevent linguistic information leakage to the top-down visual branch. We apply dropout regularization \cite{srivastava_dropout_2014} to language representation $r$ with $0.2$ probability. We use Adam optimizer \cite{kingma_adam_2014} with default parameters. We freeze the DeepLab-v3+ ResNet-101 weights. There are $32$ examples in each minibatch. We train our model for $20$ epochs on a Tesla V100 GPU with mixed precision and each epoch takes at most two hours depending on the dataset.

\noindent \textbf{Language-guided Image Colorization.}
Unless otherwise speficied, we follow the same design choices applied for the \MakeLowercase{\segmentationTask}~task. We set the number of language-conditional filters as $512$, replace the LSTM encoder with a BiLSTM encoder, and we use the first two layers of ResNet-101 trained on ImageNet as image encoder to have a similar model capacity and make a fair comparison with the previous work \cite{manjunatha_learning_2018}. We set input image width and height to $224$ in both training and validation. Thus, the low-level visual feature map has the size of $28 \times 28 \times 512$, and we don't use location features. Additionally, in our experimental analysis, we consider the same design choices with previous work \cite{manjunatha_learning_2018,zhang_colorful_2016}. %
Specifically, we use LAB color space, and our model predicts $ab$ color values for all the pixels of the input image. We perform the class re-balancing procedure to obtain class weights for weighted cross entropy objective. We use $313$ $ab$ classes present in ImageNet dataset, and encode $ab$ color values to classes by assigning them to their nearest neighbors. We use input images with a size of $224\times 224$, and output target images with a size of $56\times 56$ which is same with the previous work.

\subsection{Ablation Studies}
\label{sec:complete-ablations}

\begin{table*}[!t]
\centering
\resizebox{\textwidth}{!}{
\begin{tabular}{c|cccccc|cccccc}
\hline
\# & Top-down & Bottom-up & Depth & Layer & Visual & Textual & \textit{p@0.5} & \textit{p@0.6} & \textit{p@0.7} & \textit{p@0.8} & \textit{p@0.9} & \textit{IoU} \\ \hline \hline

\multirow{3}{0.5em}{1}
& \checkmark &  & 4 & Conv & ResNet-50 & LSTM & 66.40 & 58.59 & 49.35 & 36.01 & 13.42 & 58.06 \\
& & \checkmark & 4 & Conv & ResNet-50 & LSTM & 71.40 & 65.14 & 57.36 & 45.11 & 19.04 & 60.74 \\
& \checkmark & \checkmark & 4 & Conv & ResNet-50 & LSTM & 75.12 & 70.08 & 63.32 & 50.50 & 22.29 & 63.59 \\ \hline \hline

\multirow{3}{0.5em}{2}
& \checkmark & \checkmark & 3 & Conv & ResNet-50 & LSTM & 69.96 & 63.13 & 55.04 & 41.33 & 15.98 & 60.23 \\
& \checkmark & \checkmark & 4 & Conv & ResNet-50 & LSTM & 75.12 & 70.08 & 63.32 & 50.50 & 22.29 & 63.59 \\
& \checkmark & \checkmark & 5 & Conv & ResNet-50 & LSTM & 75.56 & 70.59 & 63.82 & 51.68 & \textbf{22.84} & 63.52 \\
\hline \hline

\multirow{2}{0.5em}{3}
& \checkmark & \checkmark & 4 & Conv & ResNet-50 & LSTM & 75.12 & 70.08 & 63.32 & 50.50 & 22.29 & 63.59 \\
& \checkmark & \checkmark & 4 & FiLM & ResNet-50 & LSTM & 71.18 & 65.14 & 57.32 & 44.66 & 18.75 & 61.12 \\
\hline \hline

\multirow{2}{0.5em}{4}
& \checkmark & \checkmark & 4 & Conv & ResNet-50 & LSTM & 75.12 & 70.08 & 63.32 & 50.50 & 22.29 & 63.59 \\
& \checkmark & \checkmark & 4 & Conv & ResNet-50 & BERT & 75.60 & 70.39 & 63.05 & 49.93 & 21.16 & 63.57 \\ \hline \hline

\multirow{2}{0.5em}{5}
& \checkmark & \checkmark & 4 & Conv & ResNet-50 & LSTM & 75.12 & 70.08 & 63.32 & 50.50 & 22.29 & 63.59 \\
& \checkmark & \checkmark & 4 & Conv & ResNet-101 & LSTM & \textbf{76.67} & \textbf{71.77} & \textbf{64.76} & \textbf{51.69} & 22.73 & \textbf{64.63} \\
\hline
\end{tabular}
}
\caption{The complete ablation studies on the UNC validation set  with \textit{p@X} and overall \textit{IoU} metrics.}
\label{table:complete-ablation}
\end{table*}

We performed additional ablation experiments on \textit{\MakeLowercase{\segmentationTask}} task in order to understand the contributions of the remaining components of our model. We share results in Table \ref{table:complete-ablation}. Each row stands for a different architectural setup. Horizontal lines separate the different ablation studies we performed, and first column denotes the ablation study group. Columns on the left determine these architectural setups. \checkmark on the \textit{Top-down} column indicates that the corresponding setup modulates top-down visual branch with language, and similarly \checkmark on the \textit{Bottom-up} column indicates that the corresponding setup modulates bottom-up visual branch with language. \textit{Depth} indicates how many layers the multi-modal encoder has. \textit{Layer} indicates the type of language-conditional layer used. \textit{Visual} and \textit{Textual} indicates which visual encoder and textual encoder used for the corresponding setup, respectively. The remaining columns stand for results.

\noindent \textbf{Network Depth (2).} We performed experiments by varying the depth size of the multi-modal encoder. We originally started with the depth size of 4. Increasing the depth size slightly increased the scores for some metrics, but more importantly, decreasing the depth size caused the model to perform worse than the bottom-up baseline. This happens because decreasing the depth size shrinks the receptive field of the network, and the model becomes less capable of drawing conclusions for the scenes that requires to be seen as a whole in order to fully understand.

\noindent \textbf{FiLM vs. Language-conditional Filters (3).} Another method for modulating language is using conditional batch normalization \cite{de_vries_modulating_2017} or its successor, FiLM layers. When we replaced language-conditional filters with FiLM layers in our model, we observed $\approx$2.5 IoU decrease. This is natural, since the FiLM layer can be thought as grouped convolution with language-conditional filters, where the number of groups is equal to number of channels / filters. %

\noindent \textbf{LSTM vs. BERT as language encoder (4).} We also experimented with BERT \cite{devlin-etal-2019-bert} as input language encoder in addition to LSTM network. We update BERT weights simultaneously with the rest of our model, where we use a smaller learning rate for BERT  ($0.00005$). We use the \textit{CLS} output embedding as our language representation $r$, than split this embedding into pieces to create language-conditional filters. Our model achieved similar quantitative results using BERT as language encoder. This points out a language encoder pre-trained on solely textual data might be sub-optimal for integrating vision and language.

\noindent \textbf{The impact of the visual backbone (5).}
We first start training our model with DeepLabv3+ ResNet-50 backbone pre-trained on Pascal VOC dataset. Then, we pre-trained a DeepLabv3+ with ResNet-101 backbone on COCO dataset by excluding the images appear on the validation and the test splits of all benchmarks similar to the previous work \cite{yu_mattnet_2018,luo2020multi,vision-language-transformer}. We only used 20 object categories present in Pascal VOC. Thus, using a more sophisticated visual backbone resulted with $\approx1$ improvement on the \textit{IoU} score.

\end{document}